\documentclass[letterpaper, 10 pt, conference]{ieeeconf}  % Comment this line out if you need a4paper

\usepackage{times}
\usepackage[numbers]{natbib}
\usepackage{multicol}
\usepackage[bookmarks=true]{hyperref}

\usepackage{lipsum}
\usepackage{comment}
\usepackage{balance}
\usepackage{graphicx}
\usepackage[table]{xcolor}
\usepackage{caption}
\usepackage{listings}
\usepackage{xcolor}

\definecolor{codegreen}{rgb}{0,0.6,0}
\definecolor{codegray}{rgb}{0.5,0.5,0.5}
\definecolor{codepurple}{rgb}{0.58,0,0.82}
\definecolor{backcolour}{rgb}{0.95,0.95,0.92}

\lstdefinestyle{mystyle}{
    backgroundcolor=\color{backcolour},
    commentstyle=\color{codegreen},
    keywordstyle=\color{magenta},
    numberstyle=\tiny\color{codegray},
    stringstyle=\color{codepurple},
    basicstyle=\ttfamily,
    breaklines=true,
    captionpos=b,
    keepspaces=true,
    numbers=left,
    numbersep=5pt,
    showstringspaces=false,
    tabsize=2
}
\usepackage{minted}
\usepackage{xcolor}
\definecolor{bg}{HTML}{f2f2ea}
\usepackage{float} % to force code snippets in place no matter what with H instead of h!
\usepackage{seqsplit}

\IEEEoverridecommandlockouts                              % This command is only needed if 
\usepackage{tabularx} 
\title{\Large \bf
Beyond URDF: The Universal Robot Description Directory for \\ Shared, Extensible, and Standardized Robot Models
}

\author{\authorblockN{Roshan Klein-Seetharaman and Daniel Rakita}
\authorblockA{Department of Computer Science, 
Yale University}
\authorblockA{\{roshan.klein-seetharaman, daniel.rakita\}@yale.edu}
}

\begin{document}

\maketitle
\thispagestyle{empty}
\pagestyle{empty}

%%%%%%%%%%%%%%%%%%%%%%%%%%%%%%%%%%%%%%%%%%%%%%%%%%%%%%%%%%%%%%%%%%%%%%%%

\begin{abstract}
Robots are typically described in software by specification files (e.g., URDF, SDF, MJCF, USD) that encode only basic kinematic, dynamic, and geometric information. As a result, downstream applications such as simulation, planning, and control must repeatedly re-derive richer data, leading to redundant computations, fragmented implementations, and limited standardization. In this work, we introduce the Universal Robot Description Directory (URDD), a modular representation that organizes derived robot information into structured, easy-to-parse JSON and YAML modules. Our open-source toolkit automatically generates URDDs from URDFs, with a Rust implementation supporting Bevy-based visualization.  Additionally, we provide a JavaScript/Three.js viewer for web-based inspection of URDDs. Experiments on multiple robot platforms show that URDDs can be generated efficiently, encapsulate substantially richer information than standard specification files, and directly enable the construction of core robotics subroutines. URDD provides a unified, extensible resource for reducing redundancy and establishing shared standards across robotics frameworks.  We conclude with a discussion on the limitations and implications of our work.

% Robots are typically described in software by specification files (e.g., URDF, SDF, MJCF, USD) that encode only basic kinematic, dynamic, and geometric information. As a result, downstream applications such as simulation, planning, and control must repeatedly derive richer data from scratch, leading to redundancy, fragmented implementations, and a lack of standardized resources. In this work, we introduce the Universal Robot Description Directory (URDD), a modular representation that organizes derived robot information into structured subdirectories, or modules, stored in easy-to-parse JSON and YAML files. Example modules include mappings between degrees of freedom and joint indices, paths between links in the kinematic chain, and the overall kinematic hierarchy supporting forward kinematics. We provide an open-source toolkit for automatically generating URDDs from URDFs, including a Rust implementation with Bevy-driven visualizations and a lightweight JavaScript/Three.js module for web-based exploration. Experiments on several robot platforms demonstrate that URDDs can be generated efficiently, store more information than standard specification files, and can be used to directly construct foundational robotics subroutines. URDD establishes a foundation for reducing redundant preprocessing, enabling shared, extensible, and standardized resources across robotics frameworks.
\end{abstract}

%DR will write

\section{Introduction}
\label{sec:introduction}

Robots are almost universally described in software by specification files (e.g., URDF, SDF, MJCF, USD) that define their structure and parameters. These files serve as the basis for downstream applications such as physics simulation, reinforcement learning, inverse kinematics, motion planning, model predictive control, and visualization.  However, these specification files typically include only minimal, raw kinematic, dynamic, and geometric information. For instance, a URDF describes connectivity between links via joints (offsets, axes, limits), inertial properties of links, and optionally references mesh files for visual and collision geometry~\cite{Quigley2009ROSAOV}.

While such files provide a useful foundation, considerably more information must be derived downstream to support effective and convenient robotics development. A straightforward example is a robot's number of degrees of freedom: although critical for many applications, it is not specified directly in a URDF. Users must either manually count non-fixed, non-mimic joints or rely on external tools to compute it. More broadly, because specification files omit much of the information required for practical use, each downstream framework is forced to re-derive richer data from scratch, leading to redundancy, fragmented implementations, and a lack of shared, standardized resources.

In this work, we propose a new representation, the Universal Robot Description Directory (URDD), which organizes a rich set of robot information into a structured collection of files, extending far beyond the minimal data captured in traditional specification formats. The URDD is composed of multiple subdirectories, called \textit{modules}, each containing derived information about the robot stored in easy-to-parse JSON and YAML files. These modules are designed to capture extensive derived information required for downstream applications, reducing redundant code and enabling shared resources across different frameworks.

\begin{comment}
\begin{figure}[t!]
    \centering
    \includegraphics[width=\columnwidth]{figures/placeholder_column_width.png}
    \caption{ }
    \vspace{-10pt}
    \label{fig:teaser}
\end{figure}    
\end{comment}

In addition to introducing the URDD representation, we provide open-source tools to automatically generate these directories for any robot. Our Rust-based implementation converts a robot’s URDF into a URDD and includes visualization capabilities, built on the Bevy game engine, that allow users to inspect derived robot information onscreen and verify processed results. We also provide an open-source JavaScript tool, built with Three.js, that enables interactive visualization of URDD outputs directly in a web browser.\footnote{\href{https://apollo-lab-yale.github.io/apollo-resources/}{https://apollo-lab-yale.github.io/apollo-resources/}}

A URDD currently contains 15 modules, which are described throughout the paper. Example modules include: the \textit{DOF Module}, which specifies the robot's number of DOFs and defines forward and inverse mappings between DOFs and joint indices; a \textit{Connections Module}, which encodes the paths between every pair of links in the kinematic chain; and a \textit{Chain Module}, which specifies the overall kinematic hierarchy of the robot, directly supporting the construction of a forward kinematics subroutine. Importantly, because the URDD is structured as a directory of sub-modules rather than a single specification file, additional modules can be incorporated over time without risking parsing errors in downstream implementations.  In addition, each module in the URDD maintains a version tag, enabling outdated modules to be easily identified for updating or verification.

We demonstrate the effectiveness of our representation and tools by evaluating the timing and outputs of the URDF-to-URDD conversion process across three robot platforms. We further show that the resulting directories not only contain substantially more information than standard specification files but can also be directly leveraged, in both Rust and Python, to construct a forward kinematics function with only simple parsing code. Finally, we conclude with a discussion of limitations, potential extensions, and broader implications.

\section{Related Works}
\label{sec:related_works}

Robot description formats serve as the foundational layer for robotics software development, enabling simulation, planning, control, and visualization across diverse platforms. The Unified Robot Description Format (URDF) remains the most widely adopted standard for specifying robot kinematic hierarchies and geometry~\cite{Quigley2009ROSAOV}. However, as robotics applications have grown in complexity, researchers have identified significant limitations in existing description formats and have proposed various extensions and alternatives.

\subsection{Robot Description Format Analysis and Extensions}

Recent work has analyzed the usage patterns and limitations of URDF in practice. Tola and Corke~\cite{Tola2023UnderstandingUAW, Tola2023UnderstandingUAAF} conducted comprehensive studies examining URDF usage across robotics applications, identifying common issues and user experience challenges. Their analysis revealed that while URDF provides essential kinematic information, it lacks many derived properties needed for practical applications.

Several researchers have proposed extensions to address URDF's limitations. Chignoli et al.~\cite{Chignoli2024URDFAEX} introduced URDF+, which extends the format to support robots with kinematic loops, a significant limitation of the original specification. Similarly, Batto et al.~\cite{Batto2025ExtendedUAAJ} developed Extended URDF to account for parallel mechanisms in robot descriptions. These works highlight the ongoing need for richer robot representations that capture more complex kinematic structures.

While these extensions address specific structural limitations of URDF, they fundamentally differ from our approach in several key ways. First, these works extend the URDF specification itself, maintaining the monolithic XML format that requires parsing and processing by each downstream application. In contrast, URDD operates orthogonally to any base specification format: it can be generated from URDF, URDF+, or other formats, then provides preprocessed, derived information in modular JSON/YAML files that eliminate redundant computation. Second, while URDF+ and Extended URDF focus on capturing more complex kinematic structures within the specification, URDD addresses the broader challenge of organizing and standardizing the derived information that applications repeatedly compute from any robot description. Our modular directory structure enables incremental extension without affecting existing parsers, whereas specification-level extensions require updates to all downstream tools. Thus, URDD complements rather than competes with these format extensions, providing a unified preprocessing layer that could benefit from richer input specifications while solving the distinct problem of redundant derivation across robotics frameworks.

\subsection{Code Generation and Preprocessing Approaches}

The concept of preprocessing robot descriptions to generate optimized code has been explored in several contexts. Frigerio et al.~\cite{Frigerio2013ADSZ, Frigerio2016RobCoGenACAC} developed domain-specific languages and code generation tools (RobCoGen) for creating efficient kinematics and dynamics implementations. Their approach generates robot-specific code in multiple programming languages, demonstrating the value of offline preprocessing for runtime performance.

Similar preprocessing strategies have been applied to motion planning and control. Astudillo et al.~\cite{Astudillo2020TowardsAOF} explored preprocessing techniques for fast nonlinear model predictive control, while motion primitive approaches have long utilized precomputed motion segments to accelerate planning algorithms. These works establish the broader principle that offline computation can significantly improve runtime performance in robotics applications.

Our work shares the preprocessing philosophy with these approaches but differs in scope and implementation strategy. While RobCoGen focuses specifically on generating executable kinematics and dynamics code for particular programming languages, URDD provides language-agnostic, structured data modules that can be consumed by any framework or programming language. Rather than generating specialized code, we precompute and organize derived information (DOF mappings, kinematic hierarchies, geometric approximations) in standardized formats that eliminate the need for repeated derivation while maintaining flexibility across different computational backends. This data-centric approach contrasts with the code generation paradigm, offering broader applicability at the cost of some runtime optimization that specialized generated code might provide.

\subsection{Simulation and Visualization Frameworks}

Modern robotics frameworks increasingly require rich geometric and dynamic information for simulation and visualization. The Gazebo simulator~\cite{Koenig2004DesignAUAX} uses SDF (Simulation Description Format) for enhanced physics simulation capabilities, while MuJoCo~\cite{Todorov2012MuJoCoAPAK} employs MJCF for efficient physics-based simulation. These specialized formats capture information beyond what URDF provides, but lack standardization across platforms.

Recent work has explored automatic generation of robot models for simulation. Lin et al.~\cite{Lin2024AutoURDFURAO} developed AutoURDF for unsupervised robot modeling from point cloud data, while various researchers have created tools for converting between different description formats~\cite{Singh2022McMujocoSAAS, Singh2024UnityARAL}. However, these conversion approaches often result in information loss and format-specific limitations.

Unlike these simulation-specific formats and conversion tools, URDD provides a simulation-agnostic intermediate representation that can serve multiple backends simultaneously. Rather than converting between incompatible formats (often with information loss), our approach generates comprehensive derived data once and makes it available in standardized, parsable formats. This feature eliminates the need for format-specific converters and reduces the coupling between robot descriptions and particular simulation engines. Our web-based and Bevy-based visualization tools demonstrate this flexibility, showing how the same URDD can drive different rendering backends without modification.

\subsection{Geometric Processing and Collision Detection}

Collision detection and geometric processing represent critical applications that require rich geometric information. Several researchers have developed specialized tools for geometric approximation and collision checking. Coumar et al.~\cite{Coumar2025FoamATS} created Foam, a tool for spherical approximation of robot geometry, while Nechyporenko et al.~\cite{Nechyporenko2025MorphItFSC} developed MorphIt for flexible spherical approximation supporting representation-driven adaptation.

These geometric processing tools typically require preprocessing of mesh data into simplified representations such as convex hulls, bounding volumes, and convex decompositions. However, this preprocessing is often performed repeatedly across different frameworks, leading to redundant computation and inconsistent results.

URDD directly addresses the redundancy problem identified in geometric processing workflows. Rather than each application independently computing convex hulls, bounding volumes, and collision-skip matrices, our mesh and link shape modules provide these geometric approximations in standardized formats that can be reused across frameworks. This design eliminates the repeated computation that tools like Foam and MorphIt perform, while ensuring consistency in geometric representations.

\section{Universal Robot Description Directory}
\label{sec:urdd_structure}

This section provides an overview of the Universal Robot Description Directory (URDD). We first discuss the overall structure, then examine several core sub-modules contained within the directory.    

\subsection{URDD Structure}

\begin{figure*}[t!]
    \includegraphics[width=\textwidth]
    {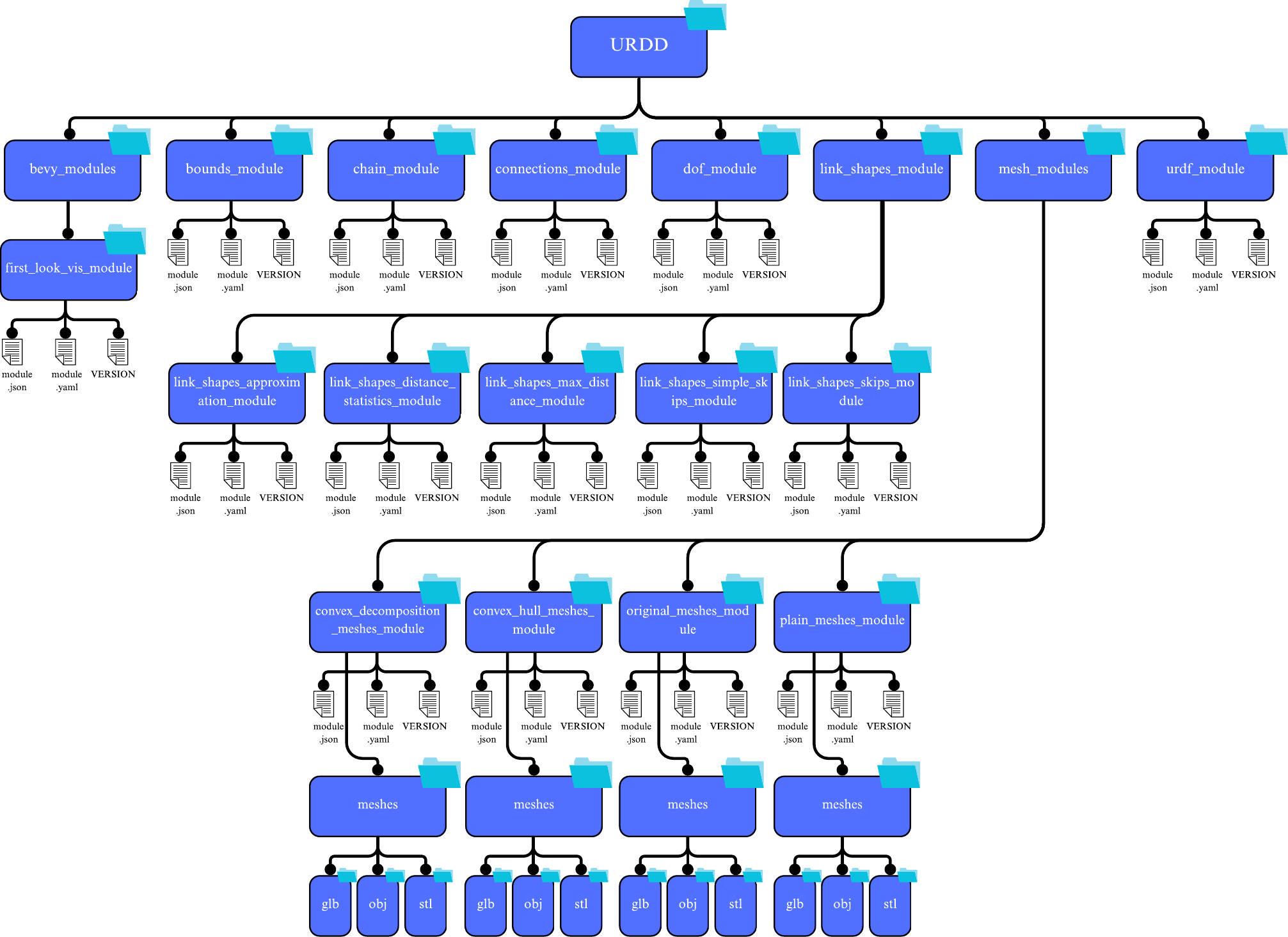}
    \caption{Structure of the Universal Robot Description Directory (URDD). The URDD organizes derived robot information into modular subdirectories stored in JSON/YAML. Core modules include kinematic structure (chain, connections, DOF mappings), joint bounds, and preprocessed geometry (meshes, convex hulls, bounding volumes). Each module is version-tagged and independently extensible, enabling richer, reusable data for planning, control, and visualization without the redundancy of re-deriving information from raw URDFs.}
    \label{fig:urdd_structure}
\end{figure*}

A URDD is organized as a hierarchical directory structure that groups together all derived robot information in a modular and extensible fashion. Figure~\ref{fig:urdd_structure} provides an overview of this structure. At the top level, the URDD contains a root directory that houses a set of subdirectories, or \emph{modules}, each responsible for encoding a particular aspect of the robot's kinematic, geometric, or dynamic properties. Every module is stored in a standardized, human-readable format (JSON or YAML), ensuring that downstream applications can easily parse and reuse the information.

The root of the URDD also maintains a metadata file that records the robot's name and versioning information. This design supports both backward compatibility and incremental extension: new modules can be incorporated without affecting existing ones, while outdated modules can be flagged and updated in isolation. In practice, this approach eliminates the brittleness often encountered when augmenting monolithic specification formats such as XML-based URDF.

The URDD separates modules into logical categories. Some modules describe structural properties (e.g., link hierarchies, kinematic chains), others capture numerical mappings (e.g., degrees of freedom indices), while others contain precomputed geometric data (e.g., convex decompositions, bounding volumes). 

% Importantly, the modular structure allows researchers and developers to only query or load the specific subdirectories required for their application, reducing overhead and encouraging interoperability across toolchains.

% In addition to the JSON and YAML text files, each module within the \texttt{mesh\_modules} subdirectory may contains a dedicated \texttt{meshes} folder. This folder stores mesh representations of robot links or subcomponents in multiple formats, including \texttt{.glb}, \texttt{.obj}, and \texttt{.stl}. By co-locating these meshes with the module data, the URDD ensures that geometric information is both immediately accessible and format-flexible for downstream tasks such as simulation, collision checking, or visualization. The module's text files reference these meshes via relative file paths, which preserves portability and avoids hard-coded dependencies on specific system directories. This design enables URDDs to be easily transferred across platforms, ensuring that both numerical and geometric resources remain synchronized and self-contained.

% Overall, the URDD structure provides a standardized, extensible foundation that moves beyond single-file specifications. By packaging robot information into a clear, modular directory, the URDD enables consistent reuse of preprocessing results, simplifies the integration of new robot models, and establishes a foundation for building shared resources across the robotics community.

\subsection{URDD Modules}
\label{sec:modules}
As noted above, each URDD is composed of a collection of modules, each encoding a distinct aspect of a robot's structure, geometry, or dynamics in lightweight JSON and YAML files.  Here, we highlight several representative modules currently included in the URDD.  For a comprehensive view of all modules, please see our documentation\footnote{\href{https://florentine-option-14b.notion.site/APOLLO-Toolbox-Documentation-9f8c4855ff3949af97c3f88d9ef04216?source=copy_link}{https://tinyurl.com/2ff8fryn}}.

\textbf{URDF Module.} This module preserves all information from the raw URDF specification but reformats it into lightweight, human-readable JSON and YAML files. By mirroring the original URDF in more accessible formats, it ensures full compatibility with existing tools while simplifying parsing and downstream use.

\textbf{DOF Module.} This module specifies the number of degrees of freedom (DOFs) of the robot and provides forward and inverse mappings between joint indices and DOF indices. Such mappings are critical for defining configuration vectors, supporting tasks like motion planning and optimization.

\textbf{Connections Module.} This module encodes the paths between all pairs of links in the kinematic tree. Each path is represented as a sequence of joints and links, enabling efficient traversal queries that support algorithms such as inverse kinematics or Jacobian construction.

\textbf{Chain Module.} This module specifies the parent–child hierarchy of the robot, listing for each link its parent joint and all associated child joints. The resulting hierarchy directly supports the construction of forward kinematics routines and other recursive computations.

\textbf{Bounds Module.} This module specifies lower and upper joint limits for each degree of freedom, enabling consistent use of configuration constraints across planning, control, and optimization routines.

\textbf{Mesh Modules.} This set of modules store original and derived mesh data for each link. Raw meshes are saved in common formats (\texttt{.glb}, \texttt{.obj}, \texttt{.stl}), alongside derived convex hulls and convex decompositions. Relative paths to these files are maintained in the module’s JSON/YAML entries, ensuring portability. This structure facilitates reuse across visualization engines, physics simulators, and collision checkers.

\textbf{Link Shapes Modules.} This set of modules provides simplified geometric approximations, such as oriented bounding boxes and bounding spheres, computed both at the link level and for each element of a convex decomposition. In addition to purely geometric data, the modules include learned representations (e.g., neural networks that approximate the robot's self-collision state~\cite{rakita2018relaxedik, rakita2021collisionik}) as well as link distance statistics (such as mean, minimum, and maximum distances) that can be leveraged by self-collision algorithms~\cite{rakita2022proxima}. The set also includes metadata identifying link pairs that can be safely skipped during self-collision checks, mirroring strategies employed in existing frameworks but stored here in a standardized, transferable format~\cite{klein2025meshpreprocessor}.

% This set of modules provides simplified geometric approximations such as oriented bounding boxes and bounding spheres, computed both at the link level and for each element of a convex decomposition. It also includes metadata identifying link pairs that can be safely skipped in self-collision checks, mirroring strategies employed in existing frameworks but stored here in a standardized, transferable format. 

% Taken together, these modules illustrate the diversity of information captured by the URDD: from basic numerical mappings to precomputed geometric properties. Importantly, the modular structure allows new capabilities to be added incrementally over time. For example, future modules could provide inertia matrices, precomputed Jacobians, or motion primitive libraries, all without disrupting existing implementations. This design positions the URDD as an extensible foundation for shared robot resources.

\section{Tools for generating and inspecting URDD}
\label{sec:tools}

While the URDD provides a standardized structure for organizing derived robot information, its practical value depends on the availability of tools that can generate and inspect these directories efficiently. To this end, we provide an open-source software suite that automates the conversion of traditional URDFs into URDDs and enables interactive visualization of their contents. These tools are designed to reduce redundant preprocessing effort, facilitate debugging, and ensure that the derived modules remain transparent and accessible to developers. 

% In this section, we describe two complementary components: (1) a converter that automatically transforms a robot’s specification into a fully populated URDD; and (2) a set of inspection tools that allow users to explore and validate URDD outputs both locally and in a web browser.

\subsection{URDF to URDD Converter}

The first tool we provide is a converter that automatically transforms a robot's URDF file into a fully populated URDD. The converter parses the minimal structural information encoded in the URDF, such as link definitions, joint types, and mesh file references, and expands it into the richer set of derived modules described in \S\ref{sec:modules}. This process includes automatically computing degrees of freedom mappings, establishing kinematic hierarchies, generating convex hulls and convex decompositions for collision geometry, and exporting meshes into multiple formats.

Our implementation is written in Rust and emphasizes both efficiency and portability. The Rust code performs the URDF parsing, generates all module files in JSON and YAML formats, and stores accompanying mesh files (\texttt{.glb}, \texttt{.obj}, \texttt{.stl}) directly within the relevant module subdirectories. Importantly, the preprocessor also includes dedicated routines for automatically generating convex hulls and convex decompositions for every shape, ensuring that geometric simplifications are available in standardized formats for downstream collision checking and proximity queries. 

Beyond batch generation, the Rust preprocessor provides an interactive, GUI-based tool, built on the Bevy game engine, that enables users to visually specify link-skip pairs for each shape type (e.g., convex hulls, oriented bounding boxes, convex decomposition elements).  This functionality streamlines the creation of self-collision matrices by allowing users to directly inspect and refine which link pairs should be excluded from collision and proximity checks.

\begin{figure}[t!]
    \centering
    \includegraphics[width=\linewidth]{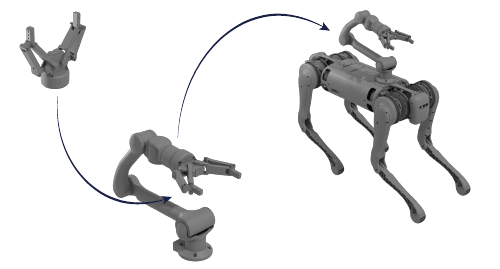}
    \caption{Our software tools enable seamless combination of multiple URDDs. In this example, a Robotiq gripper (left) is attached to a Unitree Z1 arm (center), which is then mounted onto a Unitree B1 quadruped (right). The resulting composite URDD integrates the information from all three platforms into a single, unified directory. Figure made using APOLLO Blender~\cite{messina2025apolloblender}}
    \label{fig:urdd_combination}
\end{figure}

In addition to generating URDDs for individual robots, the converter supports batch processing across entire robot repositories. This capability enables research groups and organizations to preprocess large collections of robot models at once, creating a consistent library of URDDs that can be shared and reused across projects. Each generated directory is self-contained, with relative file paths and version tags ensuring that URDDs remain portable across operating systems and computing environments.

Beyond preprocessing single robots, the converter also supports combining multiple URDDs into composite systems. For example, a gripper can be attached to the end of a manipulator arm, which in turn can be mounted on the back of a quadruped base, as illustrated in Figure~\ref{fig:urdd_combination}. These attachment points are defined using any legal joint type, including fixed, revolute, prismatic, or floating joints. This functionality allows complex, multi-robot configurations to be constructed directly from existing URDDs without requiring manual edits to the underlying URDF files.

Together, these features position the URDF-to-URDD converter as a useful tool for adopting the URDD representation, allowing existing robot descriptions to be seamlessly converted into a richer, extensible format.

\subsection{URDD Inspection Tools}

In addition to the converter, we provide a suite of inspection tools that enable developers to visualize, debug, and validate URDD outputs. These tools are designed to ensure transparency in the preprocessing stage and to simplify the process of integrating URDDs into diverse workflows. Two primary implementations are currently available.

\begin{figure}[t!]
    \centering
    \includegraphics[width=\linewidth]{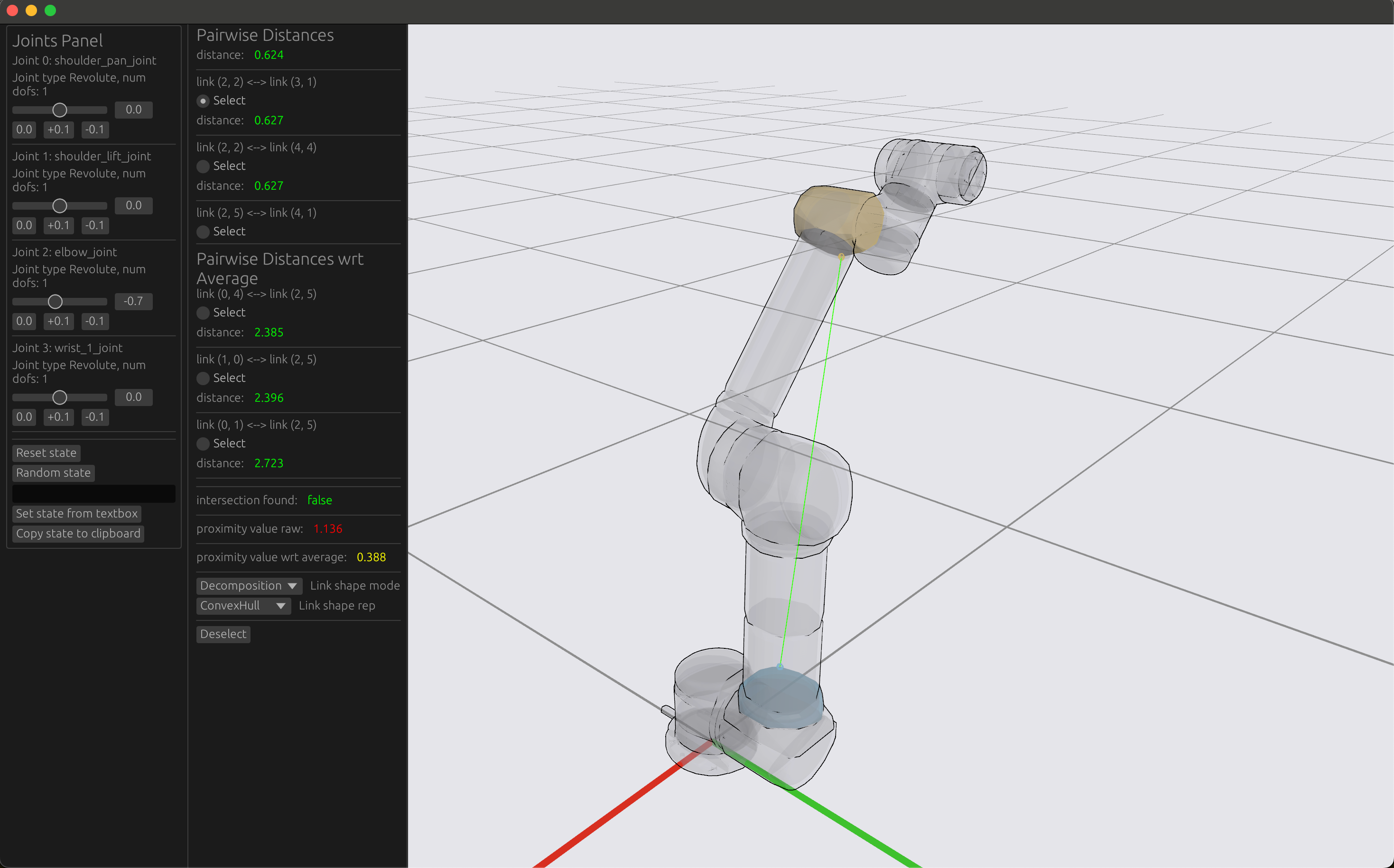}
    \caption{The Bevy-based graphics front-end powers a proximity visualization, enabling users to observe distances between pairs of link shapes. In this instance, the visualization shows the distance between two convex decomposition shapes. }
    \label{fig:proximity_vis}
\end{figure}

First, a Rust-based visualization program leverages the Bevy game engine to render robot models and their associated derived shapes in real time. Users can load any URDD and interactively inspect meshes, convex hulls, convex decompositions, and bounding volumes (seen in Figure~\ref{fig:proximity_vis}). The interface also supports toggling between different shape types, highlighting link hierarchies, and overlaying collision skip pairs, providing an intuitive means to verify the correctness of preprocessing routines. This tool is especially useful during development, as it bridges the gap between raw text-based modules and their geometric meaning.

\begin{figure}[t!]
    \centering
    \includegraphics[width=\linewidth]{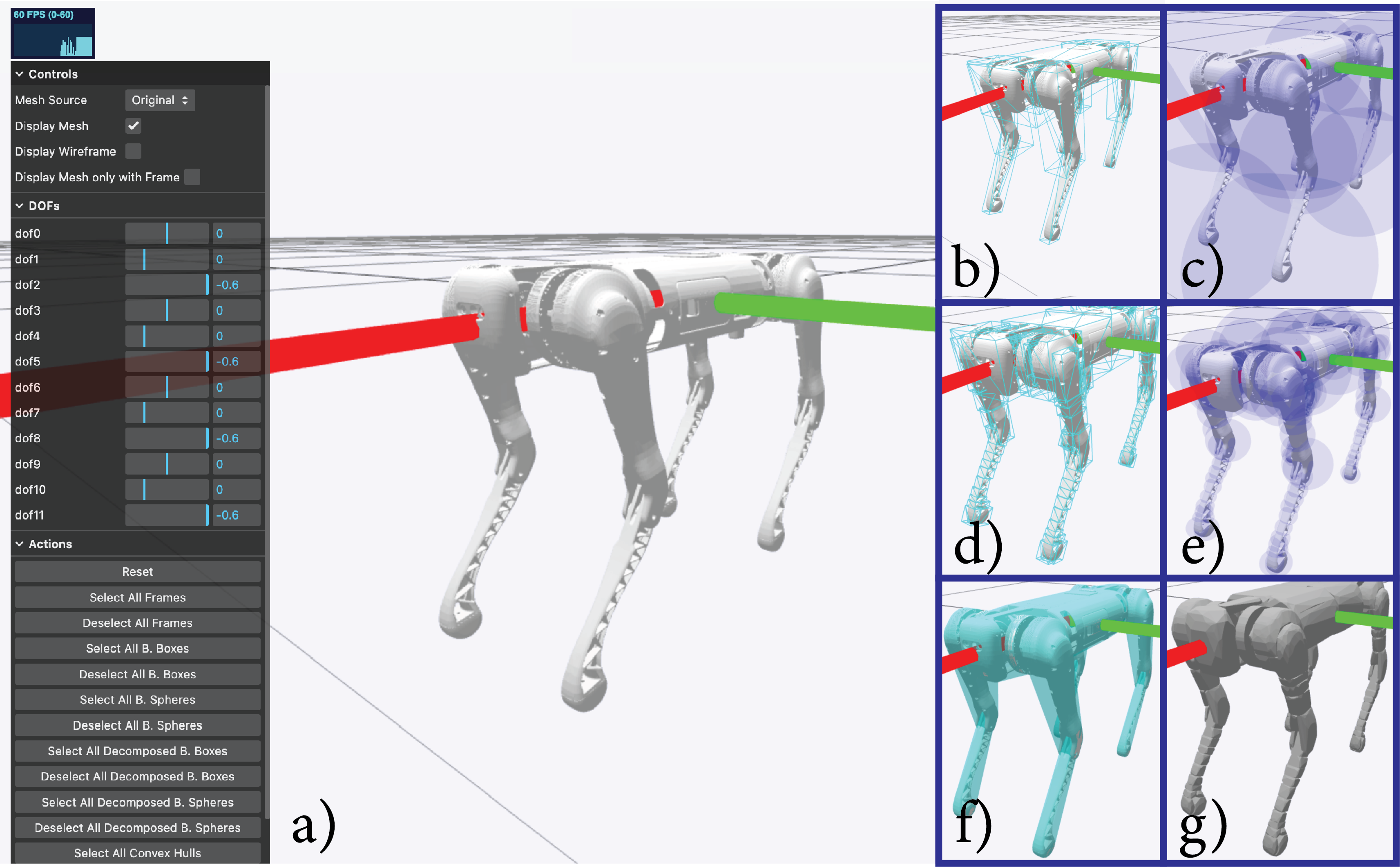}
    \caption{In-browser visualization. \textbf{a)} Overall UI and visualization of the B1 robot. \textbf{b)} Oriented bounding boxes for each link. \textbf{c)} Bounding spheres for each link. \textbf{d)} Oriented bounding boxes for each element of the convex decomposition. \textbf{e)} Bounding spheres for each element of the convex decomposition. \textbf{f)} Convex hulls. \textbf{g)} Convex decomposition for each link.}
    \label{fig:inbrowser_visualization}
\end{figure}

Second, we provide a lightweight web-based viewer implemented in JavaScript with Three.js (seen in Figure~\ref{fig:inbrowser_visualization}). This viewer reads directly from the URDD directory and renders robot models inside a browser, requiring no additional installation. By exposing URDD contents in an accessible, platform-independent environment, the web viewer makes it easy to share and validate robot resources across research groups or with collaborators who may not have access to the Rust toolchain. The viewer supports interactive features such as link highlighting, mesh type selection, and configuration-space exploration through simple slider controls.

Together, these inspection tools ensure that the URDD is not treated as a black-box data format, but rather as a transparent, verifiable resource. By making derived modules easy to visualize and validate, they reduce the likelihood of silent preprocessing errors and encourage adoption of the URDD as a shared standard for robot description.
\section{Evaluation}
\label{sec:evaluation}

To evaluate the efficacy of the Universal Robot Description Directory (URDD) and its associated tools, we conducted a series of experiments across multiple robot platforms. The goal of these experiments is to demonstrate that URDDs can be generated efficiently, that they provide a compact yet expressive representation compared to traditional URDF files, and that they can be directly applied to downstream robotics tasks with minimal additional effort. 

% Our evaluation is organized into three sub-experiments. First, we measure the runtime performance of the URDF-to-URDD conversion process, highlighting the efficiency of the Rust-based preprocessor. Second, we compare the size and structure of URDF and URDD representations, illustrating how derived information is stored in lightweight, easily parsable formats. Finally, we demonstrate how URDDs can be used to rapidly implement forward kinematics routines, showing that the modular directory structure directly supports core robotics computations.

\subsection{URDF to URDD Timing}
\label{sec:timing}

We first evaluate the efficiency of the URDF-to-URDD conversion process. Since the URDD framework is designed to eliminate redundant preprocessing by generating all necessary modules up front, it is important that this conversion step can be performed quickly, even for complex robot models. To assess this, we measured the runtime required to generate complete URDDs from a set of representative robots of varying size and complexity.  This preprocessing and data collection was done on a MacBook Air laptop equipped with an Apple M3 processor and 32GB of RAM

We run this process on five simulated robot platforms: (1) Universal Robots UR5e\footnote{\href{https://www.universal-robots.com/products/ur5e/}{https://www.universal-robots.com/products/ur5e/}}; (2) A UFactory XArm7\footnote{\href{https://www.ufactory.us/xarm}{https://www.ufactory.us/xarm}}; (3) A Unitree B1\footnote{\href{https://shop.unitree.com/products/unitree-b1}{https://shop.unitree.com/products/unitree-b1}}; (4) An Orca hand gripper\footnote{\href{https://www.orcahand.com/}{https://www.orcahand.com/}}; and (5) A Unitree H1\footnote{\href{https://www.unitree.com/h1}{https://www.unitree.com/h1}}.  The robot models outputted from this step via URDDs are all viewable in the browser visualization linked in the introduction.     

Table~\ref{tab:conversion-times} reports conversion times for five example platforms. Across these trials, the preprocessing consistently completed within seconds, with only modest variation depending on the complexity of the underlying meshes and the number of links and joints. These results indicate that the overhead of generating a URDD is negligible compared to the potential downstream savings, making it practical to preprocess large robot repositories in batch.

\begin{table}[h]
\centering
\caption{URDF-to-URDD conversion times for five example robots.  We also list the number of degrees of freedom and number of links for each robot.}
\label{tab:conversion-times}
\begin{tabular}{cccc}
\hline
\textbf{Robot} & \textbf{Num. DOFs} & \textbf{Num. Links} & \textbf{Conversion Time (s)} \\
\hline
UR5 & 6 & 11 & 25.6 \\
XArm7 & 7 & 10 & 21.3 \\
B1 & 12 & 35 & 60.0 \\
Orca hand & 17 & 55 & 90.5 \\
H1 & 19 & 25 & 69.3 \\
\hline
\end{tabular}
\end{table}

Overall, these results demonstrate that URDDs can be generated efficiently across a range of robot models, ensuring that the benefits of standardized preprocessing can be realized without imposing significant setup cost.

\subsection{URDF vs. URDD Data Size}

We next compare the information content of URDFs and URDDs. A URDF file typically encodes only the minimal structural properties of a robot (e.g., links, joints, inertial parameters), and while it may reference external mesh files, it omits many forms of derived data needed for downstream use. In contrast, a URDD explicitly stores this derived information in dedicated modules, alongside colocated meshes in multiple formats. As a result, URDDs occupy more disk space, but this increase reflects a richer and more immediately useful representation rather than redundancy.

Table~\ref{tab:data-sizes} summarizes representative sizes for the same five robots specified in \S\ref{sec:timing}, showing raw URDFs, URDFs plus referenced meshes, URDDs without meshes, and URDDs with meshes. The key observation is that URDDs capture significantly more derived information, such as DOF mappings, kinematic hierarchies, convex decompositions, and link shape approximations, while still remaining reasonably lightweight. 
% Importantly, the additional size is directly proportional to the value of the added modules, which eliminate repeated preprocessing and improve interoperability across toolchains.

\begin{table}[h]
\centering
\caption{URDF vs. URDD sizes for five example robots. URDDs store richer derived information while remaining compact.}
\label{tab:data-sizes}
\begin{tabular}{lcc|cc}
\hline
 & \textbf{URDF} & \textbf{URDF} & \textbf{URDD} & \textbf{URDD} \\
& w/o meshes & w/ meshes & w/o meshes & w/ meshes \\
 & (MB) & (MB) & (MB) & (MB) \\
\hline
UR5 & 0.013  & 6.7 & 7.5 & 31.8 \\
XArm7 & 0.017  & 2.1 & 8.0 & 17.3 \\
B1 & 0.042  & 39.6 & 13.6 & 109.8 \\
Orca hand & 0.049 & 4.1 & 43.8 & 62.4 \\
H1 & 0.028 & 33.1 & 25.8 & 112.8 \\
\hline
\end{tabular}
\end{table}

Overall, these results show that URDDs expand the amount of accessible robot information far beyond what is provided by a URDF, while maintaining compactness and portability. We argue that, in many settings, the increase in size is outweighed by the benefit of storing precomputed modules that can be directly applied in planning, control, simulation, and visualization tasks.

\subsection{Forward kinematics analysis}

Finally, we evaluate the practicality of URDDs by measuring how quickly a user can implement forward kinematics (FK), a fundamental capability in nearly all robotics applications. We tested this process using our URDD implementations in both Rust and Python. We treat FK as a ``baseline milestone'' for a robotics framework: once FK is available, other downstream capabilities such as Jacobian construction, inverse kinematics, or dynamics routines can naturally follow. Thus, the number of lines of code required to reach FK provides a useful proxy for how much preprocessing effort is pushed onto the user.

We compare our Rust and Python code to five open-source, commonly used robotics frameworks: (1) PyKDL; (2) Klampt; (3) Drake; (4) Isaac Sim; and (5) MuJoCo.

\subsubsection{Methodology} 
To obtain a fair comparison across frameworks, we determined the number of lines of code required to reach FK by back-tracing from the FK function call all the way to the initial parsing of the robot specification file. Specifically, we examined the source code of each framework and identified every line of code in the dependency chain between (1) parsing a specification file (e.g., URDF, MJCF, USD); and (2) the first callable FK function. This includes intermediate routines such as building kinematic chains, mapping DOFs to joint indices, and constructing hierarchical link structures. Importantly, our counts exclude the FK function itself and exclude generic file-parsing utilities. Only lines of code directly required along the FK dependency path are included.

\subsubsection{Results} 
In the case of URDD, both the Rust and Python implementations required \textit{no additional lines beyond parsing}. Because modules such as the chain and DOF modules already store the precomputed data needed for FK, simply loading these modules suffices to afford FK. Thus, the count of ``0'' lines for URDD in Table~\ref{tab:fk-lines} reflects the fact that \textit{all necessary logic is contained in the URDD files themselves}, leaving the user with no intermediate preprocessing burden. By contrast, frameworks such as Isaac Sim, MuJoCo, PyKDL, Klampt, and Drake must re-derive this information from raw specification files, resulting in significantly larger dependency paths.

Table~\ref{tab:fk-lines} reports the approximate number of lines of code required for each framework. As shown, URDD-based implementations reduce the preprocessing burden to zero, while other frameworks typically require hundreds or even thousands of lines of supporting code before FK becomes available.

\begin{table}[h]
\centering
\caption{Approximate lines of code required to reach forward kinematics (FK) across different frameworks. Counts are obtained by back-tracing the dependency tree from FK to the specification file.}
\label{tab:fk-lines}
\begin{tabular}{l c c}
\hline
\textbf{Framework} & \textbf{Specification Type} & \textbf{Lines to FK} \\
\hline
PyKDL              & URDF & 730  \\ % 831
Drake              & URDF & 880 \\ % 1225
Klampt             & URDF & 315 \\ % 1423
Isaac Sim          & USD  & 1892 \\ % 2034
MuJoCo             & MJCF & 3784 \\ % 6162
Custom Rust (Ours)   & URDD & 0 \\
Custom Python (Ours) & URDD & 0 \\
\hline
\end{tabular}
\end{table}

This analysis underscores the value of the URDD: by precomputing and organizing the derived modules required for FK, it collapses the dependency chain to a single step: parsing the URDD. This dramatically lowers the barrier to entry for new robot models and provides a consistent foundation for building more advanced capabilities.

\section{Discussion}
\label{sec:discussion}

In this paper, we present the Universal Robot Description Directory (URDD) representation: a modular, extensible foundation for robot description that moves well beyond the minimalism of existing formats such as URDF. By structuring derived information into lightweight, human-readable modules, the URDD reduces redundant preprocessing, enables immediate access to core data structures, and simplifies the implementation of downstream robotics functions. Our evaluation showed that URDDs can be generated efficiently, capture significantly more information than raw specification files, and reduce the barrier to entry for implementing fundamental capabilities such as forward kinematics.

A central strength of the URDD is its \emph{expandability}. Because the directory is composed of independent submodules, new modules can be added incrementally without disrupting existing workflows. For example, future modules could encode precomputed Jacobians, motion primitive libraries, or even data-driven models such as learned dynamics approximators. The modular structure also facilitates domain-specific extensions: researchers interested in surgical robotics, aerial robotics, or multi-robot systems could all add tailored modules while preserving compatibility with the broader ecosystem. In this way, the URDD serves not only as a static representation but as an evolving framework that grows with the needs of the robotics community.

\subsection{Limitations and Future Directions}

Despite its benefits, the URDD design introduces several challenges. One concern is scalability: as more modules are added, the size and complexity of each URDD may grow substantially, potentially leading to storage and parsing overhead. While our current evaluation suggests that URDDs remain lightweight, large-scale adoption across domains will require strategies to manage this growth.

A natural solution is to allow users to specify which modules are generated and stored for a particular robot. For example, a developer focused solely on kinematics may not need high-fidelity convex decompositions, while a researcher in simulation may prefer to include all geometric approximations. In addition, a clear mechanism for specifying module \emph{dependencies} would further enhance usability. Certain applications may require particular modules (e.g., Jacobian computation depends on the chain and DOF modules) and these relationships should be encoded to guide both generation and downstream use.

Finally, as the ecosystem expands, standardized documentation and validation tools will be critical for ensuring consistency across modules developed by different groups. Our current inspection tools provide a starting point, but broader community engagement will be essential to establish conventions that keep the URDD both extensible and reliable.

In summary, the URDD represents a step toward shared, standardized robot descriptions that are both rich and adaptable. By embracing modularity and expandability, it lays the groundwork for a collaborative infrastructure that can evolve alongside the diverse and growing demands of robotics research and applications.

%%%%%%%%%%%%%%%%%%%%%%%%%%%%%%%%%%%%%%%%%%%%%%%%%%%%%%%%%%%%%%%%%%%%%%%%

%%%%%%%%%%%%%%%%%%%%%%%%%%%%%%%%%%%%%%%%%%%%%%%%%%%%%%%%%%%%%%%%%%%%%%%

% \section*{ACKNOWLEDGMENT}

%%%%%%%%%%%%%%%%%%%%%%%%%%%%%%%%%%%%%%%%%%%%%%%%%%%%%%%%%%%%%%%%%%%%%%%

\bibliographystyle{plainnat}
\bibliography{refs}

% \begin{thebibliography}{99}
% \end{thebibliography}

\end{document}